\pdfoutput=1

\documentclass[11pt]{article}

\usepackage{ACL2023}

\usepackage{times}
\usepackage{latexsym}

\usepackage[T1]{fontenc}

\usepackage[utf8]{inputenc}

\usepackage{microtype}

\usepackage{inconsolata}

\usepackage{graphicx}

\usepackage{amsmath}

\usepackage{multirow}
\usepackage{xtab}

\usepackage{array}
\newcolumntype{L}[1]{>{\raggedright\let\newline\\\arraybackslash\hspace{0pt}}m{#1}}
\newcolumntype{C}[1]{>{\centering\let\newline\\\arraybackslash\hspace{0pt}}m{#1}}
\newcolumntype{R}[1]{>{\raggedleft\let\newline\\\arraybackslash\hspace{0pt}}m{#1}}

\title{Fine-tuning the SwissBERT Encoder Model for\\ Embedding Sentences and Documents}

\author{Juri Grosjean \and Jannis Vamvas\\
  Department of Computational Linguistics, University of Zurich\\
  \texttt{jurileander.grosjean@uzh.ch},~~\texttt{vamvas@cl.uzh.ch}}

\begin{document}
\maketitle
\begin{abstract}
Encoder models trained for the embedding of sentences or short documents have proven useful for tasks such as semantic search and topic modeling.
In this paper, we present a version of the SwissBERT encoder model that we specifically fine-tuned for this purpose.
SwissBERT contains language adapters for the four national languages of Switzerland – German, French, Italian, and Romansh – and has been pre-trained on a large number of news articles in those languages.
Using contrastive learning based on a subset of these articles, we trained a fine-tuned version, which we call \mbox{SentenceSwissBERT}.
Multilingual experiments on document retrieval and text classification in a Switzerland-specific setting show that SentenceSwissBERT surpasses the accuracy of the original SwissBERT model and of a comparable baseline.
The model is openly available for research use.\footnote{\url{https://huggingface.co/jgrosjean-mathesis/sentence-swissbert}}
\end{abstract}

\section{Introduction}

Sentence embeddings have become a valuable tool in natural language processing. Neural models are fed with sequence strings and convert them into embeddings, i.e. a numeric representation of the input text. These can be applied in a variety of contexts, e.g. information retrieval, semantic similarity, text classification and topic modeling.

SwissBERT \citep{vamvas2024swissbert} is a modular encoder model based on X-MOD \citep{pfeiffer-etal-2022-lifting}, which was specifically designed for multilingual representation learning.
SwissBERT has been trained via masked language modeling on more than 21 million Swiss news articles in Swiss Standard German, French, Italian, and Romansh Grischun. The model is designed for processing Switzerland-related text, e.g. for named entity recognition, part-of-speech tagging, text categorization, or word embeddings.

The aim of this work is to fine-tune the existing SwissBERT model for the embedding of sentences and short documents.
Specifically, our hypothesis is that using a contrastive learning technique such as SimCSE~\cite{gao2022simcse} to fine-tune SwissBERT will yield a model that outperforms the base model as well as generic multilingual sentence encoders in the context of processing news articles from Switzerland.

This is evaluated on two natural language processing tasks that utilize sentence embeddings, namely document retrieval and nearest-neighbor text classification, both from a monolingual and cross-lingual perspective.
Indeed, the experiments show that the fine-tuned SwissBERT, which we call SentenceSwissBERT, has a higher accuracy than baseline models.
An especially strong effect was observed for the Romansh language, with an absolute improvement in accuracy of up to 55 percentage points over the original SwissBERT model, and up to 29 percentage points over the best SentenceBERT baseline.

\begin{figure*}[h]
\centering
\setlength\fboxsep{0pt}
\setlength\fboxrule{0.25pt}
\includegraphics[width=\textwidth]{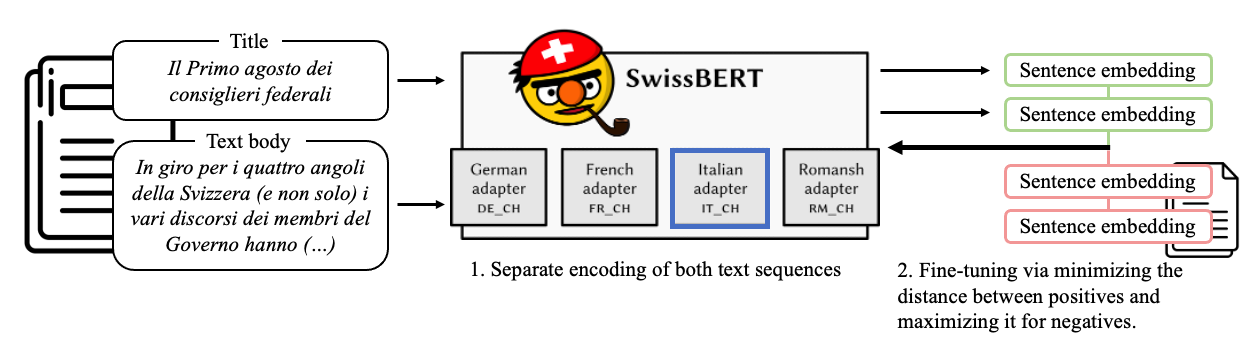}
\caption{Visualisation of the supervised SimCSE training approach.}
\label{tab:training-visual}
\end{figure*}

\section{Related Work}
\paragraph{Sentence-BERT}
This approach introduced by \citet{reimers2019sentencebert} enhances BERT and RoBERTa for generating fixed-size sentence embeddings. It investigated using the CLS-token, the mean of all output vectors (MEAN-strategy), or the max-over-time of output vectors (MAX-strategy) as sentence embeddings and found the MEAN-strategy to perform best. The method applies siamese and triplet network architectures to finetune pre-trained models, which enables them to learn high-quality sentence embeddings, e.g. for comparison via cosine similarity. The training approach entails three objective functions: classification, regression, and triplet, each with specific training structures. Data from SNLI \citep{bowman-etal-2015-large} and MultiNLI datasets \citep{williams-etal-2018-broad} was used for training.
Sentence-BERT has given rise to a family of popular open-source encoder models.\footnote{\url{https://www.sbert.net/}}

\paragraph{Multilingual Sentence Embeddings}
\label{tab:multilingual-training}
There are multiple approaches for training BERT-based encoder models for cross-lingual transfer. \citet{reimers2020making} propose utilizing knowledge distillation to enhance mono-lingual models for multilingual use.
\citet{feng2022languageagnostic} found that harnessing pre-trained language models and fine-tuning them for cross-lingual tasks yields promising results while requiring less training data than training encoder models from scratch via multilingual language data like translations.

\paragraph{Contrastive Learning}
This technique originally surged in training neural models to perform vision tasks, e.g. image recognition. However, it has also been shown to deliver promising results with NLP tasks.
The goal is for the model to learn an embedding space in which similar data is mapped closely to each other and unalike data stays far apart. For a mini-batch of $N$ sentences, where $(h_i,h_i^{+})$ represent a pair of semantically-related sequences, $h_j$ a random in-batch negative, and $\tau$ the temperature hyperparameter, the training objective looks as follows:
\begin{equation}
\label{contrastive-loss}
-\log\frac{e^{\text{cos\_sim}(h_i, h^{+}_i)/\tau}}{\sum_{j=1}^{N} e^{\text{cos\_sim}(h_i, h^{+}_j)/\tau}}
\end{equation}
Introduced by \citet{gao2022simcse}, the SimCSE (simple contrastive sentence embedding) framework has been found highly effective when used in conjunction with pre-trained language models. This technique can be applied using an unsupervised or a supervised training.

For the unsupervised approach, the sequences in the training data are matched with themselves to create positive matches, i.e. the cosine similarity between both outputs (MEAN pooling or CLS) is maximized. Thanks to the dropout masks, the embeddings of identical sequences still differ slightly.

The supervised approach uses a dataset of sentence pairs with similar meanings, and an optional third entry that is contradictory in meaning to the other two (hard negative). The similarity computation is maximized for the similar sentence pairs and minimized between the positives and the negatives.

\section{Fine-tuning}
To fine-tune SwissBERT for sentence embeddings, we opted for a (weakly) supervised SimCSE approach without hard negatives. Analogous to the original SwissBERT, Swiss news articles serve as the training data for this. The documents are split into sequence pairs, where one sequence consists of the article's title and -- if available -- its lead concatenated, while the other contains the text body (see Figure~\ref{tab:training-visual}). The title-body pairs represent $(h_i,h^{+}_i)$ in the constrastive loss training objective \ref{contrastive-loss}.

\subsection{Dataset}
The fine-tuning data consists of over 1.5 million Swiss news articles obtained through the Swissdox@LiRI database\footnote{\url{https://swissdox.linguistik.uzh.ch/}} in German, French, Italian, and Romansh (see Table~\ref{tab:training-data-stats}). All German and French articles selected from the corpus have been published between 2020 and 2023, while the Italian and Romansh media date back to 2000, because the database contains fewer articles in these languages. The news articles are pre-processed analogous to SwissBERT's original training data \citep{vamvas2024swissbert}.

\begin{table}[h]
    \begin{tabular}{L{2cm} R{2cm} R{2.5cm}}
        \textbf{Language}& \textbf{Documents}& \textbf{Tokens}\\
        \hline
        German & 760 350 & 621 107 750\\
        French & 644 416 & 567 688 406\\
        Italian & 63 666 & 35 109 282\\
        Romansh & 39 732 & 16 376 397\\
        \hline
        \textbf{Total} & \textbf{1 508 414} & \textbf{1 240 281 835}
    \end{tabular}
    \caption{Composition of the dataset used for fine-tuning SwissBERT. We report the number of documents and tokens in the four languages.}
    \label{tab:training-data-stats}
\end{table}

\subsection{Hyperparameters}
The structure of the SimCSE train script provided by \citet{gao2022simcse}\footnote{\url{https://github.com/princeton-nlp/SimCSE}} was updated and adapted according to SwissBERT, i.e. adding the X-MOD model architecture configuration as well as a language switch component, so that the model would continuously adjust its adapter according to the training data language during the training process.\\
During fine-tuning on SimCSE, we froze the language adapters and updated all the other parameters.
The training data was padded / truncated to 512 tokens, so that it fits the input limit. The model was fine-tuned in one single epoch, using a learning rate of 1e-5 and the AdamW optimizer~\cite{loshchilov2018decoupled}, a batch size of 512 and a temperature of 0.05, which has been recommended for SimCSE \citep{gao2022simcse}. We used MEAN pooling, following the findings by \citet{reimers2019sentencebert}.

\begin{table}[h]
    \begin{tabular}{L{2.5cm} L{2cm} R{2cm}}
        \textbf{Task}& \textbf{Language}& \textbf{Documents}\\
        \hline
        Document&German & 499\\
        retrieval&French & 499\\
        &Italian & 499\\
        &Romansh & 499\\
        \\
        Text classifica-&German & 4 986\\
        tion: train set&&\\
        \\
        Text classifica-&German & 1 240\\
        tion: test set&French & 1 240\\
        &Italian & 1 240\\
        &Romansh & 1 240\\
        \hline
    \end{tabular}
    \caption{Composition of the documents sourced from the \textit{20\nobreakspace Minuten} dataset \citep{zora234387} that were employed for both evaluation tasks.}
    \label{tab:20-minuten-stats}

    \begin{tabular}{L{2cm} R{2.5cm} R{2cm}}\\
        \textbf{Category} & \textbf{Train articles} & \textbf{Test articles} \\
        \hline
        accident & 244 & 60\\
        corona & 1 468 & 367\\
        economy & 768 & 192\\
        film & 247 & 61\\
        football & 627 & 156\\
        germany & 250 & 62\\
        social media & 288 & 71\\
        switzerland & 300 & 743\\
        ucraine war & 268 & 66\\
        usa & 526 & 131\\
        \hline
        \textbf{Total} & \textbf{4 986} & \textbf{1 240}\\
    \end{tabular}
    \caption{Composition of the test set of the text classification task, including the respective counts per category.}
    \label{tab:classification-stats}
\end{table}

\section{Evaluation}
We evaluate SentenceSwissBERT on two custom, Switzerland-related NLP tasks in German, French, Italian, and Romansh. It is measured against the original SwissBERT and a multilingual Sentence-BERT model that showed the strongest
performance in the given evaluation tasks.

\subsection{Dataset}
For evaluation, we make use of the \textit{20 Minuten} dataset~\citep{zora234387}, based on \textit{20 Minuten}, one of the most widely circulated German-language newspapers in Switzerland.
The articles tend to be relatively short and cover a variety of topics. Most of the documents in the dataset include a short article summary and topic tags
 
Given its format and features, the \textit{20 Minuten} dataset is especially suitable for assessing SentenceSwissBERT's performance. For the evaluation, all articles present in the \textit{20 Minuten} corpus were removed from the original fine-tuning data in all languages, so that there is no overlap.

In order to expand the evaluation to French, Italian, and Romansh, the relevant parts of the articles were machine-translated via Google Cloud API (FR, IT) and Textshuttle API (RM). Using machine translation allows for a controlled comparison across languages when evaluating, since all documents share the same structure and content. Moreover, manual annotations can be automatically projected to the other languages without a need for additional annotation. A potential downside of machine translation is that the distribution of the test data does not reflect the diversity of human-written text. Tables~\ref{tab:20-minuten-stats} and~\ref{tab:classification-stats} report statistics of the data we use for evaluation.

\subsection{Tasks}
\paragraph{Document retrieval}
For this task, the embedding of each article's summary is compared to all the articles' content embeddings and then matched by choosing the pair with the highest cosine similarity score. The performance is reported via the accuracy score, which is based on how many summaries were matched with the correct content in relation to the total number of articles processed. There is no train-test split performed for this task. It is performed monolingually (where the summary is written in the same language as the article) and cross-lingually.

\paragraph{Text Classification} 
Ten categories are manually mapped from certain topic tags in the dataset. All documents without these (or overlapping) chosen topic tags are disregarded. Then, a random train-test split with a 80/20 ratio is performed once on the remaining data for every category respectively. The exact number of files per category are displayed in Table~\ref{tab:classification-stats}. Next, the text classification is carried out utilizing a nearest neighbors approach: The text body of each test article is compared to every embedding from the training data via cosine similarity. Subsequently, the topic tag of its one nearest neighbor from the training set (highest similarity) is assigned to it.

To assess cross-lingual transfer, the training data is kept in German for the assessment of each of the four languages, while the test data is machine-translated to French, Italian and Romansh. As the categories vary in frequency, the weighted average of all categories' F1-scores is reported.

\subsection{Baseline Models}
\paragraph{SwissBERT}
While not specifically trained for this, sentence embeddings can already be extracted from the last hidden layer of the original SwissBERT encoder model via MEAN pooling. The input language is specified, just like in its newly fine-tuned version. This comparison demonstrates whether there is value in fine-tuning the model specifically for sentence embeddings.

\paragraph{Sentence-BERT}
\citet{reimers2019sentencebert} propose several multilingual sentence embedding models.\footnote{\url{https://www.sbert.net/examples/training/multilingual/README.html}} In this work, the \textit{distiluse-base-multilingual-cased-v1} model is opted for as a baseline, as it shows the strongest performance for the given evaluation tasks (see Appendix \ref{sec:appendix-b}). It has originally been trained following the multilingual knowledge distillation approach introduced in Section \ref{tab:multilingual-training}, using mUSE \citep{chidambaram-etal-2019-learning} as teacher model and a version of the multilingual Universal Sentence Encoder \citep{yang2019multilingual} as the student model. This version of Sentence-BERT supports various languages, among them French, German, and Italian, but not Romansh. Unlike with SwissBERT, the input language does not need to be specified. This model has a similar number of parameters as SwissBERT (see Table~\ref{tab:model-parameters}). However, it maps to a 512-dimensional embedding space and, hence, is computationally more efficient than SwissBERT.

The other multilingual Sentence-Transformer \textit{(paraphrase-multilingual-mpnet-base-v2)} tested is much larger (278 043 648 parameters). Although this model maps to a 768-dimensional space, analogous to SwissBERT, it performed worse in the evaluation tasks than \textit{distiluse-base-multilingual-cased-v1} (see Appendix ~\ref{sec:appendix-b}). Thus, it was disregarded.

\begin{table}[h]
    \begin{tabular}{L{2.5cm} R{2cm} R{2cm}}\\
        \textbf{Model} & \textbf{Vocabulary} & \textbf{Parameters} \\
        \hline
        Sentence-BERT & 119 547 & 135 127 808\\
        SwissBERT & 50 262 & 160 101 888\\

    \end{tabular}
    \caption{Vocabulary sizes and parameter counts of the two baseline models. The fine-tuned SentenceSwissBERT has the same size as the original model.}
    \label{tab:model-parameters}
\end{table}

\section{Results}

\paragraph{Document Retrieval}
Results for this evaluation task are reported in Table~\ref{tab:retrieval-results}. SentenceSwissBERT outperforms its base model SwissBERT, demonstrating a clear improvement compared to the original model. The largest difference is noticeable in the processing of Romansh text.

SentenceSwissBERT also obtains better results than the Sentence-BERT baseline \textit{distiluse-base-multilingual-cased}, except for two cases. Both models achieve high accuracy in both the monolingual and cross-lingual tasks. The clearest difference can be seen for German and especially Romansh, which Sentence-BERT was not trained on.

\begin{table*}[h]
    \centering
    \begin{tabular}{p{7.2cm} p{2cm} p{1.1cm} p{1.1cm} p{1.1cm} p{1.1cm}}
        \textbf{Encoder Model}& \textbf{Summary Language}& \multicolumn{2}{c}{\textbf{Article Language}}&& \\
        &&\textsc{de} & \textsc{fr} & \textsc{it} & \textsc{rm} \\
        \hline
        SwissBERT & \textsc{de} & 87.20 & 78.36 & 72.95 & 40.68 \\
        \cite{vamvas2024swissbert} & \textsc{fr} & 86.52 & 84.97 & 78.96 & 40.84 \\
        & \textsc{it} & 83.17 & 80.17 & 84.17 & 33.41 \\
        & \textsc{rm} & 46.08 & 39.10 & 43.39 & 83.17 \\
        \hline
        Sentence-BERT & \textsc{de} & 91.80 & 90.98 & \textbf{90.38} & 62.53 \\
        \cite{reimers2019sentencebert} & \textsc{fr} & 90.78 & 93.19 & 90.78 & 63.36 \\
        & \textsc{it} & 88.12 & \textbf{91.29} & 91.58 & 65.71 \\
        & \textsc{rm} & 70.59 & 73.48 & 73.55 & 73.35 \\
        \hline
        SentenceSwissBERT & \textsc{de} & \textbf{93.40}& \textbf{92.79}& 90.18& \textbf{91.58} \\
        & \textsc{fr} & \textbf{94.33}& \textbf{93.99}& \textbf{90.98}& \textbf{90.07} \\
        & \textsc{it} & \textbf{92.08}& 90.85& \textbf{92.18}& \textbf{88.50} \\
        & \textsc{rm} & \textbf{92.16}& \textbf{89.44}& \textbf{88.43}& \textbf{91.58} \\
        \hline
    \end{tabular}
    \caption{Results for the document retrieval task using the \textit{20 Minuten} dataset \cite{zora234387}. The accuracy score is reported. The best results per language pair are marked in bold print.}
    \label{tab:retrieval-results}

    \begin{tabular}{p{7.2cm} p{2cm} p{1.1cm} p{1.1cm} p{1.1cm} p{1.1cm}}\\
        \textbf{Encoder Model}& \textbf{Training Language}& \multicolumn{2}{c}{\textbf{Test Language}}&&\\
        &&\textsc{de} & \textsc{fr} & \textsc{it} & \textsc{rm} \\
        \hline
        SwissBERT \citep{vamvas2024swissbert} & \textsc{de} & 77.93& 69.62& 67.09& 43.79\\
        Sentence-BERT \citep{reimers2019sentencebert} & \textsc{de} & 77.23& 76.83& \textbf{76.90}& 65.35\\
        \hline
        SentenceSwissBERT & \textsc{de} & \textbf{78.49} & \textbf{77.18} & 76.65 & \textbf{77.20} \\
        \hline
    \end{tabular}
    \caption{Results for the nearest-neighbor classification task using the \textit{20 Minuten }dataset \citep{zora234387}. A weighted F1-score is reported and the best results are marked in bold print.}
    \label{tab:classification-results}
\end{table*}

\paragraph{Text classification}
Table~\ref{tab:classification-results} presents the results of this evaluation task. Again, SentenceSwissBERT tends to improve over the baselines, with the exception of Italian, where the Sentence-BERT model is slightly more accurate.

\section{Discussion and Conclusion}
The results confirm that contrastive learning with title–body pairs is an effective fine-tuning approach for a masked language model.
Using just a subset of 1.5 million articles from the original pre-training dataset, a clear improvement on the two sentence-level tasks has been achieved.

On the one hand, we observed an effect in monolingual tasks, e.g., by matching French summaries with French articles, or by performing nearest-neighbor topic classification of German articles using German examples.
On the other hand, we also evaluated cross-lingual variations of those tasks, and found a clear benefit in the cross-lingual setting as well, even though we did not use cross-lingual examples in our fine-tuning.
This suggests that modular deep learning with language adapters can be combined effectively with contrastive learning.

We expect that SentenceSwissBERT will be a useful model variant for other Switzerland-related tasks that require sentence or document embeddings.
For example, SentenceSwissBERT might be used for semantic search, or topic modeling based on document embeddings (e.g. BERTopic; \citealp{Grootendorst2022BERTopicNT}).
Future work could also explore whether including training data from other domains than news articles could further improve the generality of the model.

\section*{Limitations}
The SentenceSwissBERT model has been trained on news articles only. Hence, it might not perform as well on other text domains. 
Additionally, the model input during training was limited to a maximum of 512 tokens. Thus, it may not be useful for processing longer texts. Finally, we note that we used machine-translated test data for evaluation in languages other than German.


\section*{Acknowledgements}
The authors acknowledge funding by the Swiss National Science Foundation (project MUTAMUR; no.~213976).
For this publication, use was made of media data made available via Swissdox@LiRI by the Linguistic Research Infrastructure of the University of Zurich (see \url{https://t.uzh.ch/1hI} for more information).
The authors are indebted to Gerold Schneider for helpful guidance, and to Textshuttle for providing access to their Romansh machine translation API.

\bibliography{anthology,custom}
\bibliographystyle{acl_natbib}
\clearpage
\appendix
\section{Pre-training dataset media composition}
\label{sec:appendix}
    \topcaption{Media Count and Language}
    \tablefirsthead{\hline \textbf{Medium} & \textbf{Articles} & \textbf{Language} \\ \hline}
    \tablehead{\hline \textbf{Medium} & \textbf{Articles} & \textbf{Language} \\ \hline}
    \tabletail{\hline}
    \bottomcaption{Composition of the dataset used to fine-tune the SwissBERT model according to medium and language.}
    \tablelasttail{\hline}
    \small
    \begin{xtabular}{l r r}
        lematin.ch & 99 939 & \textsc{fr} \\
        24heures.ch & 73 385 & \textsc{fr} \\
        tdg.ch & 69 498 & \textsc{fr} \\
        Le Temps & 63 130 & \textsc{fr} \\
        24 heures & 62 004 & \textsc{fr} \\
        Tribune de Genève & 57 604 & \textsc{fr} \\
        blick.ch & 51 556 & \textsc{de} \\
        rsi.ch & 51 526 & \textsc{it} \\
        letemps.ch & 48 353 & \textsc{fr} \\
        rts.ch & 47 397 & \textsc{fr} \\
        cash.ch & 46 750 & \textsc{de} \\
        blick.ch & 43 178 & \textsc{fr} \\
        rtr.ch & 39 732 & \textsc{rm} \\
        srf.ch & 29 536 & \textsc{de} \\
        nzz.ch & 28 091 & \textsc{de} \\
        tagblatt.ch & 27 279 & \textsc{de} \\
        luzernerzeitung.ch & 23 855 & \textsc{de} \\
        Aargauer Zeitung / MLZ & 21 868 & \textsc{de} \\
        Neue Zürcher Zeitung & 18 408 & \textsc{de} \\
        Le Matin Dimanche & 18 352 & \textsc{fr} \\
        Thurgauer Zeitung & 17 335 & \textsc{de} \\
        Blick & 14 636 & \textsc{de} \\
        landbote.ch & 13 089 & \textsc{de} \\
        Tages-Anzeiger & 13 040 & \textsc{de} \\
        bazonline.ch & 12 709 & \textsc{de} \\
        aargauerzeitung.ch & 12 309 & \textsc{de} \\
        bernerzeitung.ch & 12 207 & \textsc{de} \\
        Zofinger Tagblatt / MLZ & 11 888 & \textsc{de} \\
        tagesanzeiger.ch & 11 612 & \textsc{de} \\
        berneroberlaender.ch & 11 603 & \textsc{de} \\
        thunertagblatt.ch & 11 581 & \textsc{de} \\
        zsz.ch & 11 517 & \textsc{de} \\
        L'Illustré & 11 231 & \textsc{fr} \\
        langenthalertagblatt.ch & 11 184 & \textsc{de} \\
        zuonline.ch & 11 120 & \textsc{de} \\
        Basler Zeitung & 10 895 & \textsc{de} \\
        derbund.ch & 10 748 & \textsc{de} \\
        schweizer-illustrierte.ch & 10 620 & \textsc{de} \\
        Zuger Zeitung & 10 557 & \textsc{de} \\
        bz - Zeitung für die Region Basel & 10 528 & \textsc{de} \\
        handelszeitung.ch & 9 790 & \textsc{de} \\
        pme.ch & 9 491 & \textsc{fr} \\
        Der Bund & 9 396 & \textsc{de} \\
        Werdenberger \& Obertoggenburger & 9 214 & \textsc{de} \\
        Der Landbote & 9 122 & \textsc{de} \\
        Zürichsee-Zeitung & 9 019 & \textsc{de} \\
        fuw.ch & 8 791 & \textsc{de} \\
        Luzerner Zeitung & 8 651 & \textsc{de} \\
        Badener Tagblatt & 8 435 & \textsc{de} \\
        Urner Zeitung & 8 284 & \textsc{de} \\
        St. Galler Tagblatt & 8 117 & \textsc{de} \\
        Wiler Zeitung & 8 003 & \textsc{de} \\
        Berner Zeitung & 7 777 & \textsc{de} \\
        Appenzeller Zeitung & 7 548 & \textsc{de} \\
        Zürcher Unterländer & 7 425 & \textsc{de} \\
        Oltner Tagblatt / MLZ & 7 420 & \textsc{de} \\
        badenertagblatt.ch & 7 140 & \textsc{de} \\
        Berner Oberländer & 7 138 & \textsc{de} \\
        Femina & 7 106 & \textsc{fr} \\
        Toggenburger Tagblatt & 7 032 & \textsc{de} \\
        Thuner Tagblatt & 6 982 & \textsc{de} \\
        solothurnerzeitung.ch & 6 120 & \textsc{de} \\
        bzbasel.ch & 5 921 & \textsc{de} \\
        RTS.ch & 5 914 & \textsc{fr} \\
        Obwaldner Zeitung & 5 854 & \textsc{de} \\
        Nidwaldner Zeitung & 5 844 & \textsc{de} \\
        TV 8 & 5 677 & \textsc{fr} \\
        Sonntagsblick & 5 606 & \textsc{de} \\
        Grenchner Tagblatt & 5 530 & \textsc{de} \\
        Solothurner Zeitung / MLZ & 5 450 & \textsc{de} \\
        BZ - Langenthaler Tagblatt & 5 277 & \textsc{de} \\
        SonntagsZeitung & 5 228 & \textsc{de} \\
        Limmattaler Zeitung / MLZ & 5 042 & \textsc{de} \\
        NZZ am Sonntag & 4 991 & \textsc{de} \\
        Finanz und Wirtschaft & 4 962 & \textsc{de} \\
        SWI swissinfo.ch & 4 855 & \textsc{it} \\
        Glückspost & 4 621 & \textsc{de} \\
        Limmattaler Zeitung & 4 513 & \textsc{de} \\
        limmattalerzeitung.ch & 4 488 & \textsc{de} \\
        rts Vidéo & 4 092 & \textsc{fr} \\
        Die Weltwoche & 4 011 & \textsc{de} \\
        Bilan & 3 979 & \textsc{fr} \\
        oltnertagblatt.ch & 3 958 & \textsc{de} \\
        grenchnertagblatt.ch & 3 857 & \textsc{de} \\
        swissinfo.ch & 3 575 & \textsc{it} \\
        www.swissinfo.ch & 3 541 & \textsc{it} \\
        swissinfo.ch & 3 525 & \textsc{fr} \\
        PME Magazine & 3 244 & \textsc{fr} \\
        illustre.ch & 3 077 & \textsc{fr} \\
        Schweizer Illustrierte & 3 068 & \textsc{de} \\
        Handelszeitung & 2 917 & \textsc{de} \\
        srf Video & 2 558 & \textsc{de} \\
        Die Wochenzeitung & 1 953 & \textsc{de} \\
        bellevue.nzz.ch & 1 919 & \textsc{de} \\
        Thalwiler Anzeiger/Sihltaler & 1 826 & \textsc{de} \\
        Zuger Presse & 1 781 & \textsc{de} \\
        HZ Insurance & 1 617 & \textsc{de} \\
        Schweizer Familie & 1 570 & \textsc{de} \\
        weltwoche.ch & 1 466 & \textsc{de} \\
        Beobachter & 1 446 & \textsc{de} \\
        Zugerbieter & 1 409 & \textsc{de} \\
        Guide TV Cinéma & 1 384 & \textsc{fr} \\
        weltwoche.de & 1 267 & \textsc{de} \\
        Tele & 1 176 & \textsc{de} \\
        Bilanz & 1 085 & \textsc{de} \\
        swissinfo.ch & 1 004 & \textsc{de} \\
        encore! & 986 & \textsc{fr} \\
        Beobachter.ch & 984 & \textsc{de} \\
        Das Magazin & 982 & \textsc{de} \\
        züritipp (Tages-Anzeiger) & 882 & \textsc{de} \\
        NZZ am Sonntag Magazin & 823 & \textsc{de} \\
        TV Star & 764 & \textsc{de} \\
        weltwoche-daily.ch & 719 & \textsc{de} \\
        bilanz.ch & 596 & \textsc{de} \\
        SWI swissinfo.ch & 587 & \textsc{fr} \\
        Streaming & 535 & \textsc{de} \\
        HZ Insurance & 529 & \textsc{fr} \\
        NZZ PRO Global & 446 & \textsc{de} \\
        Schweizer LandLiebe & 441 & \textsc{de} \\
        glueckspost.ch & 399 & \textsc{de} \\
        encore! (dt) & 274 & \textsc{de} \\
        Newsnet / 24 heures & 227 & \textsc{fr} \\
        TV Land \& Lüt & 215 & \textsc{de} \\
        NZZ Geschichte & 151 & \textsc{de} \\
        SI Sport & 143 & \textsc{de} \\
        Newsnet / Berner Zeitung & 143 & \textsc{de} \\
        Bolero & 142 & \textsc{de} \\
        boleromagazin.ch & 118 & \textsc{de} \\
        NZZ Folio & 109 & \textsc{de} \\
        beobachter.ch & 107 & \textsc{de} \\
        Aargauer Zeitung / MLZ & 91 & \textsc{fr} \\
        HZ Insurance & 77 & \textsc{it} \\
        SI Gruen & 70 & \textsc{de} \\
        L'Illustré Sport & 70 & \textsc{fr} \\
        Newsnet / Basler Zeitung & 69 & \textsc{de} \\
        Newsnet / Der Bund & 58 & \textsc{de} \\
        Bolero F & 56 & \textsc{fr} \\
        Schweiz am Wochenende & 47 & \textsc{fr} \\
        Badener Tagblatt & 34 & \textsc{fr} \\
        Schweizer Versicherung & 31 & \textsc{fr} \\
        Newsnet / Le Matin & 28 & \textsc{fr} \\
        Newsnet / Tribune de Genève & 25 & \textsc{fr} \\
        Schweizer Illustrierte Style & 23 & \textsc{it} \\
        Grenchner Tagblatt & 22 & \textsc{fr} \\
        Oltner Tagblatt / MLZ & 21 & \textsc{fr} \\
        Werdenberger \& Obertoggenburger & 21 & \textsc{it} \\
        Solothurner Zeitung / MLZ & 20 & \textsc{fr} \\
        Limmattaler Zeitung / MLZ & 20 & \textsc{fr} \\
        Finanz und Wirtschaft & 18 & \textsc{fr} \\
        NZZ Online & 16 & \textsc{de} \\
        Schweizer Versicherung & 16 & \textsc{it} \\
        TV4 & 12 & \textsc{de} \\
        Limmattaler Zeitung & 9 & \textsc{fr} \\
        rts Video & 7 & \textsc{fr} \\
        SWI swissinfo.ch & 6 & \textsc{de} \\
        Newsnet / Tages-Anzeiger & 6 & \textsc{de} \\
        Handelszeitung & 6 & \textsc{it} \\
        Berner Oberländer & 5 & \textsc{fr} \\
        Thuner Tagblatt & 5 & \textsc{fr} \\
        berneroberlaender.ch & 4 & \textsc{fr} \\
        Beobachter.ch & 4 & \textsc{it} \\
        thunertagblatt.ch & 3 & \textsc{fr} \\
        Neue Zürcher Zeitung & 3 & \textsc{it} \\
        cash.ch & 2 & \textsc{fr} \\
        Blick & 2 & \textsc{it} \\
        Berner Zeitung & 2 & \textsc{it} \\
        srf.ch & 2 & \textsc{it} \\
        weltwoche.de & 2 & \textsc{it} \\
        Blick & 1 & \textsc{fr} \\
        bernerzeitung.ch & 1 & \textsc{fr} \\
        fuw.ch & 1 & \textsc{fr} \\
        Sonntagsblick & 1 & \textsc{fr} \\
        Basler Zeitung & 1 & \textsc{fr} \\
        weltwoche.ch & 1 & \textsc{fr} \\
        weltwoche.de & 1 & \textsc{fr} \\
        srf.ch & 1 & \textsc{fr} \\
        bazonline.ch & 1 & \textsc{fr} \\
        rtr.ch & 1 & \textsc{it} \\
        derbund.ch & 1 & \textsc{it} \\
        St. Galler Tagblatt & 1 & \textsc{it} \\
        Die Weltwoche & 1 & \textsc{it} \\
        Das Magazin & 1 & \textsc{it} \\
        nzz.ch & 1 & \textsc{it} \\
        Basler Zeitung & 1 & \textsc{it} \\
        Schweiz am Sonntag / MLZ & 1 & \textsc{it} \\
        blick.ch & 1 & \textsc{it} \\
        Cash & 1 & \textsc{it} \\
        bazonline.ch & 1 & \textsc{it} \\
        \hline
    \end{xtabular}
\clearpage

\onecolumn

\section{Evaluation results of Sentence-BERT baselines}
\label{sec:appendix-b}
\begin{table}[h]
    \centering
    \begin{tabular}{p{7.2cm} p{2cm} p{1.1cm} p{1.1cm} p{1.1cm} p{1.1cm}}
        \textbf{Encoder Model}& \textbf{Summary Language}& \multicolumn{2}{c}{\textbf{Article Language}}&& \\
        &&\textsc{de} & \textsc{fr} & \textsc{it} & \textsc{rm} \\
        \hline
        \textit{paraphrase-multilingual-mpnet-base-v2} & \textsc{de} & 75.01 & 81.76 & 79.56 & 18.44 \\
        \cite{reimers2019sentencebert} & \textsc{fr} & 75.18 & 83.57 & 81.56 & 19.87 \\
        & \textsc{it} & 72.28 & 78.87 & 79.56 & 19.25 \\
        & \textsc{rm} & 53.64 & 53.91 & 57.11 & 19.44 \\
        \hline
        \textit{distiluse-base-multilingual-cased-v1} & \textsc{de} & \textbf{91.80} & \textbf{90.98} & \textbf{90.38} & \textbf{62.53} \\
        \cite{reimers2019sentencebert} & \textsc{fr} & \textbf{90.78} & \textbf{93.19} & \textbf{90.78} & \textbf{63.36} \\
        & \textsc{it} & \textbf{88.12} & \textbf{91.29} & \textbf{91.58} & \textbf{65.71} \\
        & \textsc{rm} & \textbf{70.59} & \textbf{73.48} & \textbf{73.55} & \textbf{73.35} \\
        \hline
    \end{tabular}
    \caption{Results for the document retrieval task using two multilingual Sentence-BERT models. The accuracy score is reported. The best results per language pair are marked in bold print.}

    \begin{tabular}{p{7.2cm} p{2cm} p{1.1cm} p{1.1cm} p{1.1cm} p{1.1cm}}\\
        \textbf{Encoder Model}& \textbf{Training Language}& \multicolumn{2}{c}{\textbf{Test Language}}&&\\
        &&\textsc{de} & \textsc{fr} & \textsc{it} & \textsc{rm} \\
        \hline
        \textit{paraphrase-multilingual-mpnet-base-v2} & \textsc{de} & 75.42& 75.64& 73.88& 39.38\\
        \citep{reimers2019sentencebert} &&&&&\\
        \hline
        \textit{distiluse-base-multilingual-cased-v1} & \textsc{de} & \textbf{77.23}& \textbf{76.83}& \textbf{76.90}& \textbf{65.35}\\
        \citep{reimers2019sentencebert} &&&&&\\
        \hline
    \end{tabular}
    \caption{Results for the nearest-neighbor classification task using the two multilingual Sentence-BERT models. A weighted F1-score is reported and the best results are marked in bold print.}
\end{table}

\end{document}